\documentclass[letterpaper, 10 pt, conference]{IEEE_styles/ieeeconf}  

\IEEEoverridecommandlockouts %only required if we need to use the \thanks command
\usepackage{0a_preamble}
\usepackage{pifont}

\usepackage{eso-pic}

\newcommand{\IEEEacceptednotice}{%
\fontsize{6}{7}\selectfont
\copyright\ 2026 IEEE. Personal use of this material is permitted.
Permission from IEEE must be obtained for all other uses, in any current or future media,
including reprinting/republishing this material for advertising or promotional purposes,
creating new collective works, for resale or redistribution to servers or lists, or reuse of
any copyrighted component of this work in other works.}

\begin{document}

%add the macros
%%%%%%%%%%%%%%%%%%%%%%%%%%%%%%%%%%%%%%%%%%%%%%%%%%%%%%%%%%%%%%%%%%%%%%%%%%%%
%%%%%%%%%%%%%%%%%%%%%%%%%%%%%%%%%%%%%%%%%%%%%%%%%%%%%%%%%%%%%%%%%%%%%%%%%%%%
% Project Descriptions/Names
%%%%%%%%%%%%%%%%%%%%%%%%%%%%%%%%%%%%%%%%%%%%%%%%%%%%%%%%%%%%%%%%%%%%%%%%%%%%
%%%%%%%%%%%%%%%%%%%%%%%%%%%%%%%%%%%%%%%%%%%%%%%%%%%%%%%%%%%%%%%%%%%%%%%%%%%%

\newcommand\ProjectName{fARfetch\xspace}

%%%%%%%%%%%%%%%%%%%%%%%%%%%%%%%%%%%%%%%%%%%%%%%%%%%%%%%%%%%%%%%%%%%%%%%%%%%%
%%%%%%%%%%%%%%%%%%%%%%%%%%%%%%%%%%%%%%%%%%%%%%%%%%%%%%%%%%%%%%%%%%%%%%%%%%%%
% Common Terms
%%%%%%%%%%%%%%%%%%%%%%%%%%%%%%%%%%%%%%%%%%%%%%%%%%%%%%%%%%%%%%%%%%%%%%%%%%%%
%%%%%%%%%%%%%%%%%%%%%%%%%%%%%%%%%%%%%%%%%%%%%%%%%%%%%%%%%%%%%%%%%%%%%%%%%%%%
%maps
\newcommand{\globalMap}{$m_{\textrm{G}}$}
\newcommand{\localMap}{$m_{\textrm{L}}$}
\newcommand{\localMapGF}{$\hat{m}_{\textrm{L}}$}

\newcommand\iwr{TI-IWR1843}
\newcommand\dca{TI-DCA1000}
\newcommand{\Lidar}{Lidar}
\newcommand{\lidar}{lidar}
\newcommand{\radar}{radar}
\newcommand{\Radar}{Radar}
\newcommand{\blue}[1]{{\color{black}#1}}

%%%%%%%%%%%%%%%%%%%%%%%%%%%%%%%%%%%%%%%%%%%%%%%%%%%%%%%%%%%%%%%%%%%%%%%%%%%%
%%%%%%%%%%%%%%%%%%%%%%%%%%%%%%%%%%%%%%%%%%%%%%%%%%%%%%%%%%%%%%%%%%%%%%%%%%%%
% Localization/Filtering Terms
%%%%%%%%%%%%%%%%%%%%%%%%%%%%%%%%%%%%%%%%%%%%%%%%%%%%%%%%%%%%%%%%%%%%%%%%%%%%
%%%%%%%%%%%%%%%%%%%%%%%%%%%%%%%%%%%%%%%%%%%%%%%%%%%%%%%%%%%%%%%%%%%%%%%%%%%%
%kalman filter
\newcommand{\KalmanStateMatrix}{\textrm{\textbf{s}}}

%IMU Output
\newcommand{\accX}{a_{\textrm{x}}}
\newcommand{\accY}{a_{\textrm{y}}}
\newcommand{\accZ}{a_{\textrm{z}}}
\newcommand{\wPitch}{\omega_\Theta}
\newcommand{\wRoll}{\omega_\Phi}
\newcommand{\wYaw}{\omega_\Psi}

%IMU biases
\newcommand{\wYawBias}{\omega_{\textrm{bias}}}

%initial positions/headings
\newcommand{\poseInitial}{\textrm{\textbf{P}}_{\textrm{0}}}
\newcommand{\poseTransInitial}{\textrm{\textbf{p}}_{\textrm{0}}}
\newcommand{\poseHeadInitial}{\Psi_{\textrm{0}}}

%estimated positions/headings
\newcommand{\poseEst}[1]{\hat{\textrm{\textbf{P}}}_{#1}}
\newcommand{\poseTransEst}[1]{\hat{\textrm{\textbf{p}}}_{#1}}
\newcommand{\poseHeadEst}[1]{\hat{\Psi}_{#1}}

%ground truth positions/headings
\newcommand{\poseGT}[1]{{\textrm{\textbf{p}}_{#1}}}
\newcommand{\poseTransGT}[1]{\textrm{\textbf{p}}_{#1}}
\newcommand{\poseHeadGT}[1]{\Psi_{#1}}

%position
\newcommand{\pose}{\hat{\textrm{\textbf{p}}}}
\newcommand{\poseX}{x}
\newcommand{\poseY}{y}

%velocity
\newcommand{\vel}{\textrm{\textbf{v}}}
\newcommand{\velX}{v_{\textrm{x}}}
\newcommand{\velY}{v_{\textrm{y}}}

%orientation
\newcommand{\pitch}{\Theta}
\newcommand{\roll}{\Phi}
\newcommand{\yaw}{\Psi}

%ugv buffer radius
\newcommand{\bufferUGV}{\textrm{R}_{\textrm{UGV}}}

%%%%%%%%%%%%%%%%%%%%%%%%%%%%%%%%%%%%%%%%%%%%%%%%%%%%%%%%%%%%%%%%%%%%%%%%%%%%
%%%%%%%%%%%%%%%%%%%%%%%%%%%%%%%%%%%%%%%%%%%%%%%%%%%%%%%%%%%%%%%%%%%%%%%%%%%%
% Paragraph Labels
%%%%%%%%%%%%%%%%%%%%%%%%%%%%%%%%%%%%%%%%%%%%%%%%%%%%%%%%%%%%%%%%%%%%%%%%%%%%
%%%%%%%%%%%%%%%%%%%%%%%%%%%%%%%%%%%%%%%%%%%%%%%%%%%%%%%%%%%%%%%%%%%%%%%%%%%%
\newcommand{\paragraphLabel}[1]{\noindent\textbf{#1}}
\newcommand{\emphasizedLabel}[1]{\noindent\textit{#1}}
\newcommand{\keyStep}[2]{\noindent\textbf{Step \circled{#1}: #2}}
\newcommand{\softwarePackage}[1]{\texttt{#1}}

%%%%%%%%%%%%%%%%%%%%%%%%%%%%%%%%%%%%%%%%%%%%%%%%%%%%%%%%%%%%%%%%%%%%%%%%%%%%
%%%%%%%%%%%%%%%%%%%%%%%%%%%%%%%%%%%%%%%%%%%%%%%%%%%%%%%%%%%%%%%%%%%%%%%%%%%%
% figure, section, eq, and table Labels
%%%%%%%%%%%%%%%%%%%%%%%%%%%%%%%%%%%%%%%%%%%%%%%%%%%%%%%%%%%%%%%%%%%%%%%%%%%%
%%%%%%%%%%%%%%%%%%%%%%%%%%%%%%%%%%%%%%%%%%%%%%%%%%%%%%%%%%%%%%%%%%%%%%%%%%%%
\newcommand{\sect}[1]{Sec.~{#1}}
\newcommand{\fig}[1]{Fig.~{#1}}
\newcommand{\eq}[1]{eq.~{#1}}
\newcommand{\tbl}[1]{Table~{#1}}

%%%%%%%%%%%%%%%%%%%%%%%%%%%%%%%%%%%%%%%%%%%%%%%%%%%%%%%%%%%%%%%%%%%%%%%%%%%%
%%%%%%%%%%%%%%%%%%%%%%%%%%%%%%%%%%%%%%%%%%%%%%%%%%%%%%%%%%%%%%%%%%%%%%%%%%%%
%Units
%%%%%%%%%%%%%%%%%%%%%%%%%%%%%%%%%%%%%%%%%%%%%%%%%%%%%%%%%%%%%%%%%%%%%%%%%%%%
%%%%%%%%%%%%%%%%%%%%%%%%%%%%%%%%%%%%%%%%%%%%%%%%%%%%%%%%%%%%%%%%%%%%%%%%%%%%
\newcommand{\dBsm}[1]{{#1}\thinspace{dBsm}}
\newcommand{\dBsmSq}[1]{{#1}\thinspace{$\textrm{dBsm}^2$}}
\newcommand{\dB}[1]{{#1}\thinspace{dB}}
\newcommand{\dBm}[1]{{#1}\thinspace{dBm}}
%times
\newcommand{\nsec}[1]{{#1}\thinspace{ns}}
\newcommand{\usec}[1]{{#1}\thinspace{$\upmu$s}}
\newcommand{\msec}[1]{{#1}\thinspace{ms}}
\newcommand{\seconds}[1]{{#1}\thinspace{s}}
%frequencies
\newcommand{\GHz}[1]{{#1}\thinspace{GHz}}
\newcommand{\MHz}[1]{{#1}\thinspace{MHz}}
\newcommand{\kHz}[1]{{#1}\thinspace{kHz}}
\newcommand{\Hz}[1]{{#1}\thinspace{Hz}}
%Data Rates
\newcommand{\MSps}[1]{{#1}\thinspace{MSps}}
\newcommand{\MBps}[1]{{#1}\thinspace{MBps}}

%slopes
\newcommand{\MHzPerus}[1]{{#1}\thinspace{MHz/$\upmu$s}}
%distances
\newcommand{\m}[1]{{#1}\thinspace{m}}
\newcommand{\mSq}[1]{{#1}\thinspace{$\textrm{m}^2$}}
\newcommand{\cm}[1]{{#1}\thinspace{cm}}
\newcommand{\mm}[1]{{#1}\thinspace{mm}}
%velocities
\newcommand{\mPers}[1]{{#1}\thinspace{m/s}}

%angles
\newcommand{\degrees}[1]{{#1}$^{\circ}$}
\newcommand{\radians}[1]{{#1}\thinspace{rad}}
\newcommand{\radiansSq}[1]{${#1\textrm{rad}}^{2}$}

%mass
\newcommand{\grams}[1]{{#1}\thinspace{g}}
\newcommand{\kilograms}[1]{{#1}\thinspace{kg}}

%power
\newcommand{\Watts}[1]{{#1}\thinspace{W}}

%constants
\newcommand{\LightSpeed}{\textrm{c}}

%%%%%%%%%%%%%%%%%%%%%%%%%%%%%%%%%%%%%%%%%%%%%%%%%%%%%%%%%%%%%%%%%%%%%%%%%%%%
%%%%%%%%%%%%%%%%%%%%%%%%%%%%%%%%%%%%%%%%%%%%%%%%%%%%%%%%%%%%%%%%%%%%%%%%%%%%
% Radar Parameters,Performance Specificeations, detections
%%%%%%%%%%%%%%%%%%%%%%%%%%%%%%%%%%%%%%%%%%%%%%%%%%%%%%%%%%%%%%%%%%%%%%%%%%%%
%%%%%%%%%%%%%%%%%%%%%%%%%%%%%%%%%%%%%%%%%%%%%%%%%%%%%%%%%%%%%%%%%%%%%%%%%%%%
%detections
\newcommand{\detection}{\textrm{\textbf{d}}}
\newcommand{\pointCloud}[1]{\mathcal{C}_{\textrm{#1}}}

\newcommand{\detectionX}{d_{\textrm{x}}}
\newcommand{\detectionY}{d_{\textrm{y}}}
\newcommand{\detectionZ}{d_{\textrm{z}}}
\newcommand{\detectionV}{d_{\textrm{v}}}
\newcommand{\detectionP}{p_{\textrm{det}}}

%GNN nodes
\newcommand{\node}{\textrm{\textbf{n}}}

%% Radar Signals
\newcommand{\TxSignal}{s_{\textrm{TX}}(t)}
\newcommand{\RxSignal}{s_{\textrm{RX}}(t)}
\newcommand{\RxSignalConj}{s_{\textrm{RX}}^{*}(t)}
\newcommand{\IFSignal}{s_{\textrm{IF}}(t)}

%% Radar Configuration
\newcommand{\NumChirpsPerFrame}{N_{\textrm{chirps}}}
\newcommand{\FrameDuration}{T_{\textrm{frame}}}
\newcommand{\ChirpDuration}{T_{\textrm{chirp}}}
\newcommand{\ChirpFreqStart}{f_{c}}
\newcommand{\ChirpWavelength}{\lambda}
\newcommand{\ChirpSlope}{S}
\newcommand{\ChirpSlopeVic}{S_{\textrm{vic}}}
\newcommand{\ChirpSlopeAtt}{S_{\textrm{atk}}}
\newcommand{\ChirpBW}{B}
\newcommand{\FreqSamp}{f_{\textrm{samp}}}
\newcommand{\NSamps}{N_{\textrm{Samp}}}
\newcommand{\NChirps}{N_{\textrm{Chirps}}}
\newcommand{\KRx}{K_{\textrm{Rx}}}
\newcommand{\KTx}{K_{\textrm{Tx}}}

% Performance Specifications
\newcommand{\RangeRes}{d_{\textrm{res}}}
\newcommand{\RangeMax}{d_{\textrm{max}}}
\newcommand{\RangeMin}{d_{\textrm{min}}}
\newcommand{\VelocityRes}{v_{\textrm{res}}}
\newcommand{\VelocityMax}{v_{\textrm{max}}}
\newcommand{\AngleRes}{\theta_{\textrm{res}}}

%radar target
\newcommand{\TargetRange}{d_{\textrm{obj}}}
\newcommand{\TargetVelocity}{v_{\textrm{obj}}}
\newcommand{\TargetAngle}{\theta_{\textrm{obj}}}

% Intermediate terms (time delays, phase shifts, IF frequencies)
\newcommand{\tDelay}{t_{\textrm{d}}}
\newcommand{\TargetPhaseShift}{\phi_{\textrm{obj}}}
\newcommand{\DopplerShift}{\phi_{\textrm{doppler}}}
\newcommand{\FreqIF}{f_{\textrm{IF}}}
%%%%%%%%%%%%%%%%%%%%%%%%%%%%%%%%%%%%%%%%%%%%%%%%%%%%%%%%%%%%%%%%%%%%%%%%%%%%
%%%%%%%%%%%%%%%%%%%%%%%%%%%%%%%%%%%%%%%%%%%%%%%%%%%%%%%%%%%%%%%%%%%%%%%%%%%%
% RadNav Parameters
%%%%%%%%%%%%%%%%%%%%%%%%%%%%%%%%%%%%%%%%%%%%%%%%%%%%%%%%%%%%%%%%%%%%%%%%%%%%
%%%%%%%%%%%%%%%%%%%%%%%%%%%%%%%%%%%%%%%%%%%%%%%%%%%%%%%%%%%%%%%%%%%%%%%%%%%%

%% Test sites
\newcommand{\Wilkinson}{\ding{182}}
\newcommand{\NorthVICON}{\ding{183}}
\newcommand{\NorthBasement}{\ding{184}}
\newcommand{\CPSL}{\ding{185}}

%% Ground vehicle parameters
\newcommand{\UGVRadius}{R_{UGV}}
%%%%%%%%%%%%%%%%%%%%%%%%%%%%%%%%%%%%%%%%%%%%%%%%%%%%%%%%%%%%%%%%%%%%%%%%%%%%
%%%%%%%%%%%%%%%%%%%%%%%%%%%%%%%%%%%%%%%%%%%%%%%%%%%%%%%%%%%%%%%%%%%%%%%%%%%%
% Comments
%%%%%%%%%%%%%%%%%%%%%%%%%%%%%%%%%%%%%%%%%%%%%%%%%%%%%%%%%%%%%%%%%%%%%%%%%%%%
%%%%%%%%%%%%%%%%%%%%%%%%%%%%%%%%%%%%%%%%%%%%%%%%%%%%%%%%%%%%%%%%%%%%%%%%%%%%%create flags to show comments/todos
\newif\ifShowComments
\newif\ifShowToDos

%set flags
\ShowCommentstrue
\ShowToDostrue

%david comments
\newcommand\david[1]{
    \ifShowComments\textcolor{blue}{[DH: #1]} \fi%
}

%shaocheng comments
\newcommand\shaocheng[1]{
    \ifShowComments\textcolor{purple}{[SL: #1]} \fi%
}

\newenvironment{myitemize}
  {\begin{list}{$\bullet$}
    {
      \setlength{\leftmargin}{0.15in}
      \setlength{\itemsep}{0.1\baselineskip} % Space between list items
      \setlength{\parsep}{0pt}               % Space between paragraphs inside items
      \setlength{\topsep}{0pt}               % Space above and below the list
      \setlength{\partopsep}{0pt}            % Extra space when list starts a paragraph
    }
  }
  {\end{list}}

\title{\LARGE \bf 
\ProjectName: Enabling Collocated AR-HRC in Large Visually Diverse Environments with VLM-Driven AR Content Adaptation}

%create flags to show comments/todos
\newif\ifAnonymize

%set flags
\Anonymizefalse

%add title if Anonymize is false
\ifAnonymize

\else
    \author{Christian Fronk, Hanting Ye, David Hunt, Miroslav Pajic, and Maria Gorlatova  %
    % \thanks{*These authors contributed equally to this work.}
    % \thanks{This work is sponsored in part by the ONR under the agreement N00014-23-1-2206, AFOSR FA9550-19-1-0169 Award, NSF CNS-1652544 and CNS-2211944 awards, and the National AI Institute for Edge Computing Leveraging Next Generation Wireless Networks, Grant CNS-2112562.}% <-this % stops a space
    \thanks{The authors are with the Department of Electrical and Computer
    Engineering, Duke University, Durham, NC 27708 USA (e-mail:
    \{christian.fronk, hanting.ye, david.hunt, miroslav.pajic, maria.gorlatova\}@duke.edu).}
    }
\fi

\maketitle

\AddToShipoutPictureFG*{%
  \AtPageLowerLeft{%
    \put(54,18){%
      \parbox{\dimexpr\paperwidth-108pt\relax}{\IEEEacceptednotice}%
    }%
  }%
}

%from Amir Template
\thispagestyle{empty}
\pagestyle{empty}

\begin{abstract}
% Augmented Reality (AR) can improve collocated human–robot collaboration by making robot state and intent visible and enabling intuitive control, yet large, visually diverse environments like the outdoors challenge both interaction and content legibility, especially at long distances and beyond visual line of sight. We present \ProjectName, an AR-HRC system that integrates (i) shared semantic environment mapping across an AR headset and robot that visualizes detected landmarks in AR to support landmark-grounded go-to commands, (ii) a context-aware world-in-miniature representation of the shared environment for fine-grained path authoring, and (iii) vision-language-model driven AR view management that jointly adapts virtual content color, size, and orientation to maintain legibility in large visually diverse environments. We implement \ProjectName with a Meta Quest 3 headset and Unitree Go2 quadruped robot, and conduct a within-subjects user study (N=13) on a real-world large-scale (30.5~m) outdoor inspection task. \ProjectName yielded significantly faster completion times than a non-AR baseline (66\%) and significantly lower workload in mental demand (-43\%), temporal demand (-34\%), and frustration (-66\%). A custom legibility survey indicated \ProjectName effectively maintained virtual content legibility in the large outdoor environment.

Augmented Reality (AR) can enhance collocated human–robot collaboration by communicating robot state and enabling intuitive control, but large, visually diverse environments challenge robot control and content legibility, particularly at long distances and beyond visual line of sight. We present \ProjectName, an AR-HRC system designed to support collocated robot control in large, visually diverse outdoor environments by integrating (i) shared semantic environment mapping to visualize landmarks and support landmark-grounded go-to commands, (ii) a context-aware world-in-miniature for path authoring, and (iii) vision-language-model driven view management that jointly adapts content color, size, and orientation to maintain legibility. We implement \ProjectName on a Meta Quest 3 and Unitree Go2 and evaluate it in a within-subjects study (N=13, 30.5~m) in a large outdoor environment. \ProjectName reduced completion time by 66\% and workload across mental (-43\%), temporal (-34\%), and frustration (-66\%) demands, while effectively maintaining content legibility.
\end{abstract}

% NASA–TLX
% for the Meta Quest~3
% To our knowledge, \ProjectName is the first AR-HRC system to support beyond-VLOS control in large environments that combines semantic landmarking with WIM-based path authoring at scale (30.5~m) and VLM-driven adaptive view management.

\section{Introduction}
\label{sec:introduction}
% Augmented reality (AR) superimposes virtual content onto the real world, allowing information to be intuitively situated to guide users through tasks such as navigation~\cite{arbasicnavigation}. 
Augmented reality (AR) virtual overlays have shown strong potential in human-robot collaboration (HRC) to enable more flexible interaction by conveying a robot’s state and intent~\cite{RaffikICAECA2023MRCobotsSurvey,SuzukiCHIARHRCSurve2022,Maccio_ROMAN_2023_Show_Robot_Intent}, such as during inspection tasks~\cite{TVCG_2014_AR_Microdrones_outdoor_inspection} or for teleoperation~\cite{IROS_2021_MR_Telepresence_Step_Into_World_No_Control}. Beyond unidirectional communication, AR also supports bidirectional interaction, enabling in-situ control of robotic collaborators. Such control has been applied across diverse platforms, including drones~\cite{Chen2021PinpointFlyCHI} and quadrupeds~\cite{ICRA_2024_MR_Robot_drag_and_drop_quadruped}. Here, users can issue commands by manipulating virtual content or directing robots toward real-world landmarks~\cite{CHI_2025_SafeSpect_AR_Drone_Building_Inspection}. Other approaches allow not only destination assignment but also path authoring, enabling trajectory modification~\cite{IJSC_AR_Robot_Arm_Control_And_PathPlanning,ISMAR_2022_DroneARchery,MMAR_2023_Robot_Arm_AR_Path_Planning} or programming behaviors for later execution~\cite{IROS_2018_Robot_Arm_AR_Programming}. 

%~\cite{RaffikICAECA2023MRCobotsSurvey,GreenIJARSHRCSurvey2008,SuzukiCHIARHRCSurve2022,ismarhrc,anticipatoryarm,staticvdynamicviz,multipleuserssinglerobot,SongVR2025VirtualUltrasound,roboticassembly,Maccio_ROMAN_2023_Show_Robot_Intent}

%such as during robotic ultrasounds~\cite{SongVR2025VirtualUltrasound} and assembly tasks~\cite{roboticassembly}.  Applications span both indoor~\cite{roboticassembly,Chen2021PinpointFlyCHI} and outdoor~\cite{TVCG_2014_AR_Microdrones_outdoor_inspection,IROS_2021_Outdoor_AR_Telepresence,IROS_2021_MR_Telepresence_Step_Into_World_No_Control} settings, highlighting AR’s versatility for HRC scenarios.

%Beyond unidirectional communication, AR also supports bidirectional interaction, enabling in-situ control of robotic collaborators. Such control has been applied across diverse platforms, including drones~\cite{Chen2021PinpointFlyCHI} and quadrupeds~\cite{ICRA_2024_MR_Robot_drag_and_drop_quadruped}. With these systems, users can issue commands by manipulating virtual content or directing robots toward real-world landmarks~\cite{CHI_2025_SafeSpect_AR_Drone_Building_Inspection}. Other approaches allow not only destination assignment but also path authoring, enabling either real-time trajectory modification~\cite{IJSC_AR_Robot_Arm_Control_And_PathPlanning,ISMAR_2022_DroneARchery,MMAR_2023_Robot_Arm_AR_Path_Planning,ISMAR_Adjunct_2023_AR_Drone_Path_Authoring} or programming of behaviors for later execution~\cite{IROS_2018_Robot_Arm_AR_Programming}.

\begin{figure}[]
\centering
\vspace{-0.0cm}
\includegraphics[width=.8\columnwidth]
{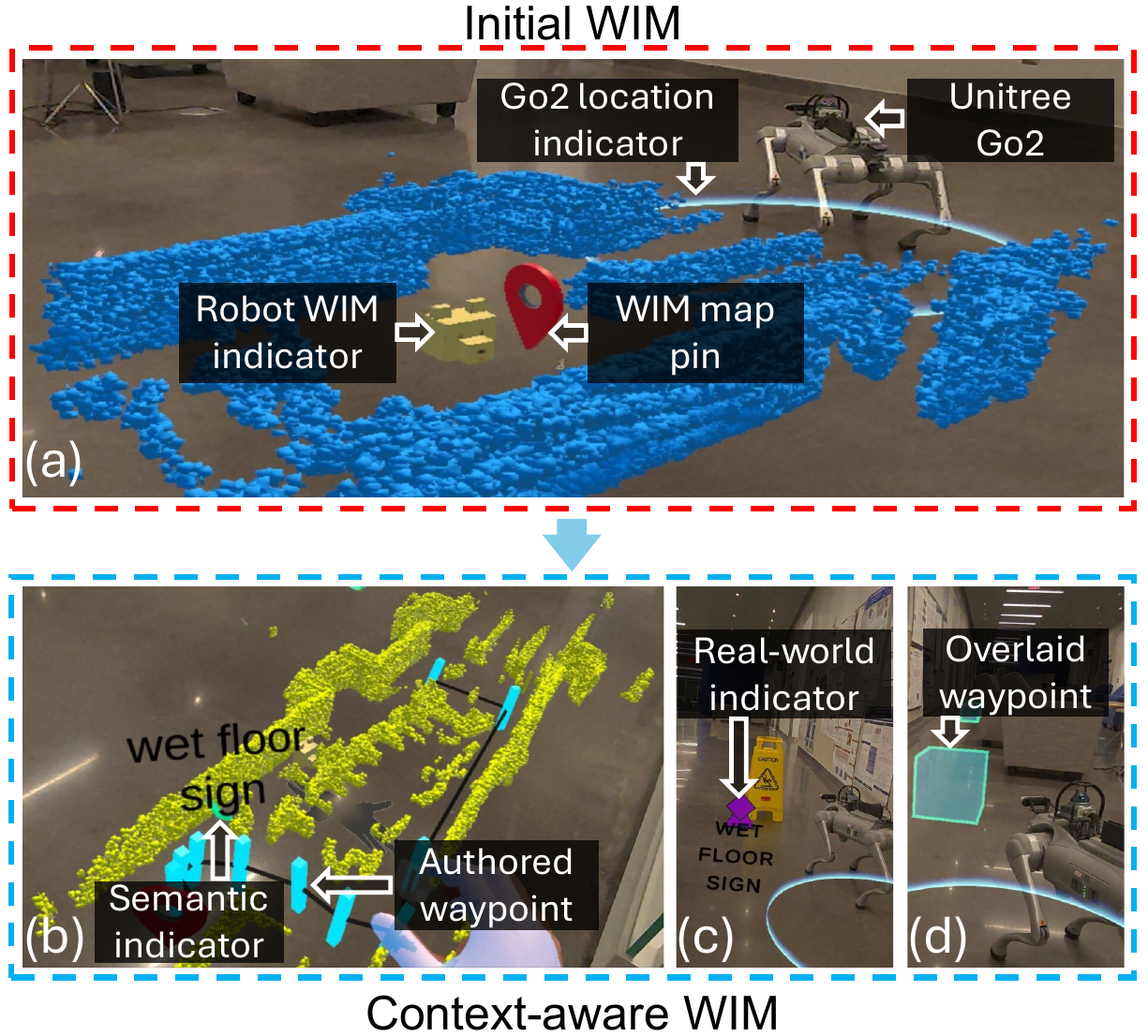}
% \vspace{-0.6cm}
\caption{%\ProjectName WIM usage. (a) Initial WIM at collaboration onset, with robot WIM indicator and map pin; in the real-world view, a translucent ring marks Go2’s location. (b) User specifies a destination and authors waypoints to route Go2 around a semantic landmark; WIM color is adapted by VLM for improved legibility. (c) Close-up of landmark with VLM-adapted indicator. (d) Go2 follows WIM-authored path; authored waypoints are overlaid in the real environment to visualize the path.
\ProjectName WIM usage. (a) Initial WIM with robot WIM indicator and map pin; in the real-world view, a translucent ring marks Go2’s location. (b) User sets a destination and authors waypoints around a semantic landmark; VLM adaptation improves WIM legibility. (c) Landmark close-up with VLM-adapted indicator. (d) Go2 follows WIM-authored path; authored waypoints are visualized in the real environment.}
\vspace{-0.5cm}
\label{wim_control_fig}
\end{figure}

However, virtual control content becomes increasingly difficult to place and manipulate in large environments. At greater distances, content appears at reduced resolution in the AR head-mounted display (HMD), and users struggle to situate it meaningfully in the real world (e.g., placing a destination marker) because their perception of the depth, scale, and relative position of distant virtual objects is diminished~\cite{Appl_Percep_2021_Est_Dist_In_AR,VR_2024_Transitive_Percep_In_AR}. 
% Interfaces that rely on physical landmarks also lose effectiveness in large spaces, as selecting faraway targets with an AR HMD has been shown to be challenging~\cite{TVCG_2022_Pointing_In_Collab_Outdoor_AR}.
World-in-miniature (WIM) interfaces \cite{ICRA_2024_MR_Robot_drag_and_drop_quadruped,ICRA_2021_ARROCH_AR_Robot_Collab_Indoors_TabletBased} have been proposed to support effective robot control in large spaces like the outdoors. 
% Existing method propose using 
However, when a user's visual line-of-sight (VLOS) is obstructed, existing WIM interfaces become less interpretable, as they provide only simple structural map visualizations as in Fig.~\ref{wim_control_fig}(a) and users lack awareness of clear semantic landmarks in the collaborator’s surrounding environment; moreover, semantic landmarks detected by the AR headset are not shared with the robot. They also do not enable user path authoring, making it difficult to achieve precise navigation with respect to landmarks. %Fundamentally, this limitation reflects a disconnect between the world models of the robot and the user: while the robot builds a geometric \textit{and} semantic understanding through its sensors, this knowledge is not directly visualized in the user’s physical surroundings. 
Therefore, a key challenge is to \emph{enable robust collocated AR control and path authoring that remains effective in large environments, preserving usability at greater robot distances and when VLOS is lost by incorporating shared semantic mapping from both the robot and AR device.}

%Therefore, a key challenge we face is \emph{how to enable collocated control and path authoring within AR while maintaining functionality in large environments, even when users are at a greater distance to the robot and VLOS to the robot cannot be guaranteed.}

% More fundamentally, the robot collaborator's surrounding environment has already been captured by its onboard sensors and is known to the robot in most cases. This limitation naturally arises from the disconnect between the robot’s internal world model and the human user’s perception of the environment in the current HRC context: \emph{while the robot builds a geometric and semantic understanding through its sensors, this knowledge is not directly visualized in the user’s physical surroundings.} 

In this paper, we introduce \ProjectName, an AR system for the Meta Quest 3 that enables collocated AR-HRC in large, visually diverse environments where robot proximity and VLOS is not guaranteed.  \ProjectName constructs a \emph{context-aware WIM} of the shared environment by labeling real-world landmarks detected through both the AR headset and the robot. These landmarks are then embedded both directly in the AR view of the environment as in Fig.~\ref{wim_control_fig}(c) and within a 3D WIM as seen in Fig.~\ref{wim_control_fig}(b). This allows users to issue contextually meaningful commands, such as navigating to a specific landmark or navigating around an obstacle, and to author precise, context-aware paths at scale, even when the robot is distant or occluded. To further support spatial understanding and control, \ProjectName also visualizes authored paths in the context-aware WIM at their corresponding real-world locations as in Fig.~\ref{wim_control_fig}(d), maintaining a consistent connection between the miniature representation and the physical environment. \ProjectName also implements pointing-style ``Go-To" destination commands using these semantic landmarks to allow users to quickly issue commands when path authoring is not required. 

To support effective human–robot collaboration, virtual content must remain consistently legible, as \emph{loss of legibility can compromise both user safety and task performance when elements conveying robot intent or enabling control are difficult to perceive}~\cite{perceptionmanip}. In large environments, legibility depends on how well content adapts to factors such as user distance, viewing angle, and background appearance, which jointly influence how easily content can be perceived. These challenges are further amplified in outdoor settings, where dynamic and visually complex backgrounds can significantly degrade legibility. Consequently, an intelligent method that considers both content and scene context in tandem is needed to reason about adaptations in large operational environments and enable adaptive AR view management to ensure consistent content legibility. To this end, \ProjectName integrates a scene-aware, VLM-driven view management approach that \textit{jointly adapts} the color, size, and orientation of virtual content in response to the surrounding scene. This facilitates unified, context-sensitive adaptations at runtime, maintaining legibility across varied environmental conditions.

We evaluate \ProjectName in a within-subjects user study in which participants using a Meta Quest 3 collaborated with a Unitree Go2 quadruped in a real-world inspection task in a large outdoor environment. Compared to a non-AR baseline,  \ProjectName significantly improved completion times (66\% faster) and significantly reduced perceived workload during the task. In summary, our main contributions are:

%In addition, \ProjectName provides persistent virtual indicators showing the robot’s current location and intended trajectory, allowing users to easily track its movements and intentions.

\vspace{-0.03cm}
\begin{myitemize}
\item We present \ProjectName, an AR-HRC system for collocated robot control in large environments where robot proximity cannot be assumed, enabling destination and fine-grained path commands at large distances (30.5~m) through shared sensing between headset and robot.
\item We support beyond-VLOS collocated operation in \ProjectName through shared semantic mapping between robot and user, embedding labeled landmarks into the real-world view and a context-aware WIM to support command issuing and path authoring at scale.
\item We present a VLM-driven, scene-aware AR view management technique that jointly adapts content color, scale, and orientation to ensure legibility across visually diverse environments such as outdoors.
\end{myitemize}

The remainder of this paper is organized as follows: Sec.~\ref{sec:related_work} reviews related work, followed by \ProjectName's system design in Sec.~\ref{sec:system_design}. We present our user study in Sec.~\ref{sec:user_study}, and the results in Sec.~\ref{sec:results}. We conclude the paper in Sec.~\ref{sec:conclusion}. 

% \vspace{-0.4cm}
\section{Related Work}
\label{sec:related_work}
% \noindent\textbf{Robot Control Through AR World-In-Miniature.}
% % AR has been widely applied in HRC to improve task efficiency~\cite{ismarhrc,staticvdynamicviz,robotinteractingvirtual,IkedaHRI2024ProgramAR,cozmo,anticipatoryarm} and enhance understanding of robot intentions and internal states. 
% Beyond one-way communication, AR has supported in-situ robot control across drones~\cite{ISIC_2016_AR_Drone_Object_Structure_Inspection,HRI_2019_Drone_Teleop_Indoors,HRI_2018_Teleop_Indoors_Analytics_Comparison,CHI_2025_SafeSpect_AR_Drone_Building_Inspection,Chen2021PinpointFlyCHI}, robotic arms~\cite{IkedaHRI2024ProgramAR,ChanHRI2022HumanRobotArmManufacturingCollab,IJSC_AR_Robot_Arm_Control_And_PathPlanning}, and ground robots~\cite{ICRA_2021_ARROCH_AR_Robot_Collab_Indoors_TabletBased,ICRA_2024_MR_Robot_drag_and_drop_quadruped}. Some of these systems introduce world-in-miniature (WIM) interfaces for visualization and control, but typically focus on simple destination commands rather than semantically grounded or context-aware path authoring~\cite{IJSC_AR_Robot_Arm_Control_And_PathPlanning,ISMAR_2022_DroneARchery,MMAR_2023_Robot_Arm_AR_Path_Planning,ISMAR_Adjunct_2023_AR_Drone_Path_Authoring}.
\noindent\textbf{AR Robot Control.}
Most AR-HRC systems that enable robot control assume robot proximity or continuous visual line-of-sight (VLOS) to the robot. Many require external controllers such as an Xbox controller~\cite{HRI_2019_Drone_Teleop_Indoors,HRI_2018_Teleop_Indoors_Analytics_Comparison} when the robot is distant or restrict operation to scenarios where the robot remains in view~\cite{Chen2021PinpointFlyCHI,IJSC_AR_Robot_Arm_Control_And_PathPlanning}. These assumptions limit applicability in large spaces like outdoors where robots often move beyond VLOS. To address such cases, some approaches visualize the robot’s planned trajectory when not in VLOS or provide structural world-in-miniature (WIM) interfaces for robot control~\cite{ICRA_2024_MR_Robot_drag_and_drop_quadruped,ICRA_2021_ARROCH_AR_Robot_Collab_Indoors_TabletBased}. However, a limitation of prior WIM-based approaches is that they typically support only destination selection and lack both semantic context and path authoring. This constrains users to issuing simple commands without the ability to design trajectories that reflect the structure of their environment based on the landmarks within it. Moreover, the robot’s environmental understanding is not fully integrated back into the user’s physical view as spatial AR augmentation. Prior methods are also primarily evaluated in computer-based simulations. \ProjectName addresses these gaps by leveraging shared sensing between the AR headset and robot to construct a shared semantic understanding of the environment, embedding landmarks detected by both the robot and headset directly into the real-world AR view and a context-aware WIM generated from robot sensor data, enabling users to issue commands relative to meaningful environmental features. Beyond this, \ProjectName extends WIM interaction to support precise path authoring, allowing users to construct context-aware trajectories at scale. We also evaluate \ProjectName in a real-world user study with a physical robot in a representative large-scale outdoor inspection task. 
% However, these methods either lack contextual information about the environment or do not support user-authored paths, making precise navigation difficult.  
% A limitation of prior WIM-based approaches for robot control is that they typically support only destination selection and lack both semantic context and the ability to author paths. This constrains users to issuing simple commands without the ability to design trajectories that reflect the structure of their environment. \ProjectName addresses this gap by embedding semantic landmarks detected by both the Quest and the robot directly into the WIM, enabling users to issue commands grounded in meaningful environmental features. Beyond this, \ProjectName extends WIM interaction to support precise path authoring, allowing users to construct context-aware trajectories at scale. We now describe the WIM generation and interaction process in detail.
%\vspace{-0.1cm}

\noindent\textbf{View Management in AR-HRC.}
Prior AR view management methods primarily target predefined, static real-world objects~\cite{HCI_2023_BlendMR,CHI_2022_ScalAR,UIST_2021_SemanticAdapt, Satkowski2022CeilingorFloorPlacement, xiu2025viddar}, focusing on repositioning virtual overlays to preserve real-world visibility. However, in AR-HRC scenarios, the legibility of virtual content conveying robot intent or control is equally critical, as poor legibility can lead to confusion and reduced task efficiency~\cite{perceptionmanip}. Generalizing prior approaches to AR-HRC is challenging due to frequent user and robot motion and dynamic, potentially large environments, which invalidate assumptions of static spatial configurations. Moreover, content often appears against varying textures, distances, and viewing angles, where perceptual clarity depends jointly on color, scale, and orientation. Adjusting these properties independently can produce conflicts; for example, reorientation may alter contrast conditions established by a prior color adjustment, highlighting the need for a method that accounts for content and scene properties in combination. Inspired by recent work demonstrating that VLMs enable rich AR scene reasoning~\cite{UIST_2025_AdjustAR, xiu2025viddar}, we introduce a VLM-driven view management strategy for large, dynamic AR-HRC environments. By leveraging a VLM’s ability to reason over a scene, our approach jointly adapts intrinsic visual properties of content at runtime, providing unified adjustments that maintain legibility across diverse large-scale environments.
\section{System Design}
\label{sec:system_design}
This section introduces \ProjectName, an AR-HRC system tailored for large-scale visually diverse environments like the outdoors where proximity to the robot and continuous VLOS cannot be assumed. The system integrates three core capabilities: (1) shared semantic environment mapping that detects and labels real-world landmarks which can be used for destination commands, (2) a world-in-miniature (WIM) representation that embeds these landmarks to support context-aware destination commands and path authoring with path waypoints also overlaid on the real-world, and (3) a VLM-driven view management module that jointly adapts the \emph{color}, \emph{size}, and \emph{orientation} of virtual content to maintain legibility in visually diverse settings. fARfetch emphasizes shared sensing between the robot and user, with objects detected by either the robot or headset being inserted into a shared semantic map and visualized in AR, allowing the robot and headset's understanding of space to collaboratively inform user control of the robot. Together, these components enable robust collocated human-robot collaboration entirely within AR, even at long distances and beyond VLOS.
\vspace{-0.1cm}
\subsection{System Implementation}
\begin{figure}[t]
\centering
\vspace{0.2cm}
\includegraphics[width=\columnwidth]
{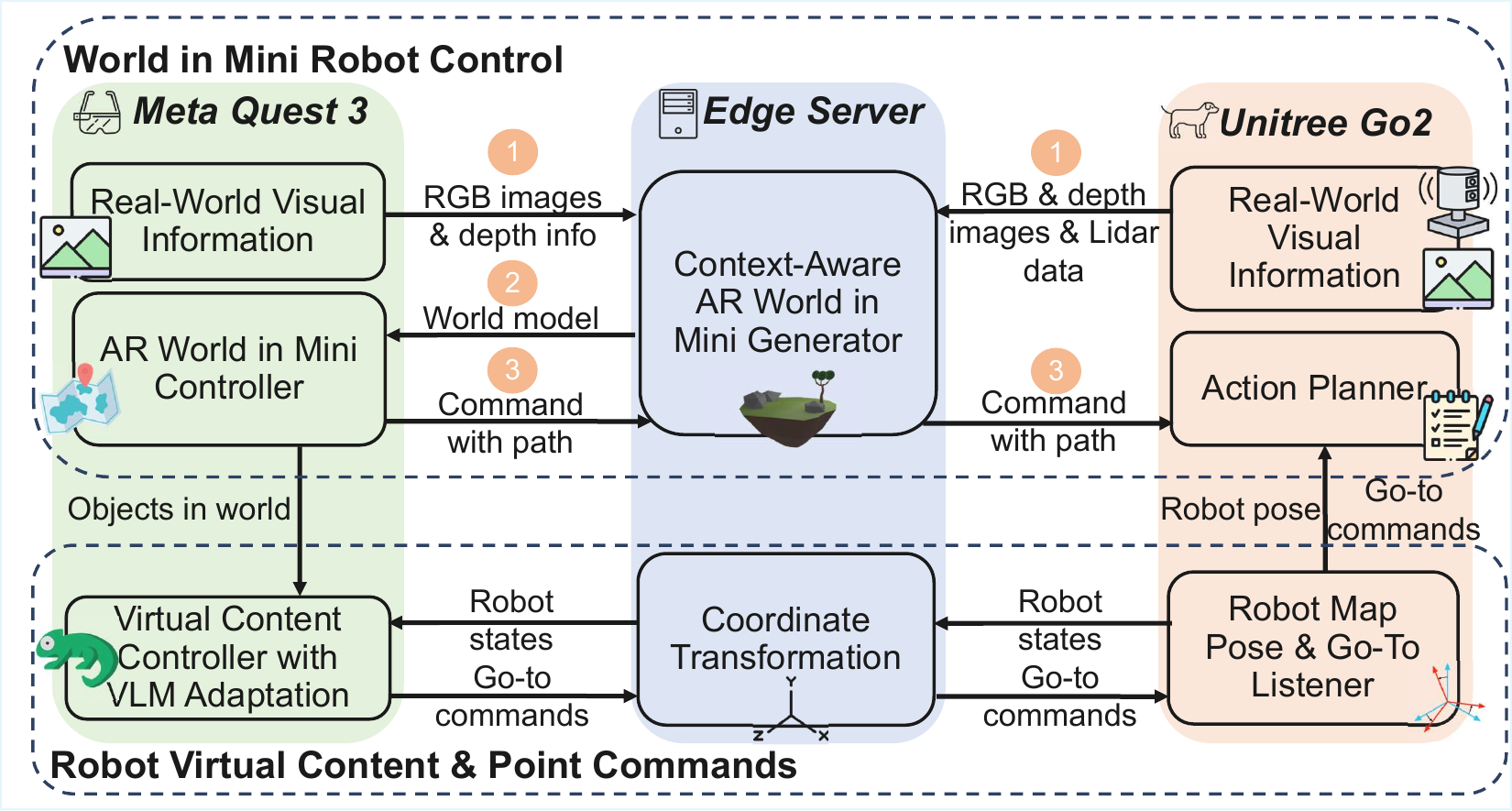}
\vspace{-0.5cm}%-.5
\caption{\ProjectName system diagram. }
\vspace{-0.5cm}
\label{system_diagram}
\end{figure}

\ProjectName consists of 3 components: an AR headset, a robot, and an edge server. We use the Meta Quest 3 as the AR headset, which runs a Unity 2022.3.10f1 application that we developed that enables user control of the robot. The robot collaborator is a Unitree Go2 quadruped running ROS 2 Foxy~\cite{ros_foxy}. It runs SLAM toolbox~\cite{slam_toolbox} and Nav2~\cite{nav2}, along with custom nodes that we created. The edge server, equipped with an NVIDIA RTX 3060 GPU, hosts a FastAPI-based~\cite{fastapi} Python API we developed to facilitate communication between the Quest and the Go2. Our server acts as a local compute resource for offloading more demanding tasks such as WIM generation. To define a transformation between the Quest world frame and Go2 map frame, we use an ArUco marker~\cite{arucomarker} as a shared calibration target. During initialization, both the Quest and Go2 observe the same marker and estimate its pose in their respective frames. Let $T^{Q}_{M}$ be the marker pose in the Quest world frame $Q$, and let $T^{G}_{M}$ be the marker pose in the Go2 map frame $G$. Because both poses correspond to the same physical marker, the rigid transform from the Go2 map frame to the Quest world frame is computed as $T^{Q}_{G} = T^{Q}_{M}(T^{G}_{M})^{-1}$. The edge server uses this transform to express robot poses and robot-detected landmarks in the Quest frame for AR visualization, and uses its inverse to convert AR-authored goals and waypoints into the Go2 map frame for navigation. A common feature of AR-HRC is virtual content that helps maintain user awareness of the robot’s position; to support this, \ProjectName employs a translucent ring to indicate the robot’s location, visible in Fig.~\ref{wim_control_fig}(a), (c), and (d). The OpenAI API for GPT-4o~\cite{gpt4osystemcard} is used for our VLM-driven adaptation. The overall architecture of \ProjectName is shown in Fig.~\ref{system_diagram}. We now introduce the main features of \ProjectName.

\subsection{World-In-Miniature (WIM) and Go-To Robot Control}

In this subsection, we introduce how we generate a context-aware WIM, and our designed robot commands.

% \subsubsection{World In Mini (WIM) Robot Control}
% A limitation of prior WIM-based approaches for robot control is that they typically support only destination selection and lack both semantic context and the ability to author paths. This constrains users to issuing simple commands without the ability to design trajectories that reflect the structure of their environment. \ProjectName addresses this gap by embedding semantic landmarks detected by both the Quest and the robot directly into the WIM, enabling users to issue commands grounded in meaningful environmental features. Beyond this, \ProjectName extends WIM interaction to support precise path authoring, allowing users to construct context-aware trajectories at scale. We now describe the WIM generation and interaction process in detail.

\subsubsection{Context-Aware WIM Generation}
A generated WIM combines the headset’s and robot’s semantic understanding of the environment with the robot’s structural map of that same environment. The Quest and Go2 each stream RGB images paired with depth data, which the edge server processes through the context-aware WIM generator, as seen in Fig.~\ref{system_diagram}. For each image, objects are detected using Grounding DINO~\cite{grounding_dino}, chosen for its generalizable, open-set recognition capabilities. Detected bounding boxes are then refined through instance segmentation with the Segment Anything Model~\cite{SAM}, allowing more accurate depth estimation. Using instance masks together with camera intrinsics, the 3D positions of objects are computed in the device’s global reference frame. To avoid cluttering the WIM with objects that are not useful for robot control, detected objects are filtered using a task-configurable vocabulary that specifies which objects are relevant in the current environment. Each retained landmark is stored in a shared dictionary, accessible to the Quest and Go2, along with its class label, 3D location, and cropped appearance. This forms \ProjectName's shared semantic map. To avoid duplicates, retained landmarks are compared against existing ones of the same class using spatial proximity and visual similarity computed via CLIP~\cite{CLIP} embeddings of each object's appearance.

In parallel, the Go2 contributes structural environment information by streaming point cloud data from its Livox Mid-360 lidar. Every three scans are aggregated as a point cloud $P_{i}$, transformed into the robot map frame, and sent to the server. The server concatenates each cloud into a global cloud $P_{O} = P_{1} + P_{2} + \dots + P_{i}$. The WIM generator voxelizes the accumulated cloud to produce a 3D mesh representing the explored environment as in Fig.~\ref{wim_control_fig}(a). The server combines the structural mesh and the semantic map by embedding labeled indicators for each retained landmark and sends the result to the AR WIM controller. A 3D indicator for the robot is also added to the WIM, and moves within it as the WIM controller receives pose updates from the robot. The result is a semantically enriched WIM that fuses geometric structure with task-relevant contextual information, as seen in Fig.~\ref{wim_control_fig}(b).

% \begin{figure}[t]
% \centering
% \vspace{-0.0cm}
% \includegraphics[width=\columnwidth]
% {figures/WIMControlFig.png}
% \vspace{-0.6cm}
% \caption{\ProjectName WIM usage. (a): Initial WIM showing the Go2 indicator and map marker at collaboration onset. (b): The user specifies a destination and adjusts waypoints to route the Go2 around a landmark (a wet-floor sign) representing a potential hazard. (c): Close-up of the landmark with a VLM-adapted indicator. (d): The Go2 following the WIM-authored waypoints overlaid on the real environment.}
% \vspace{-0.5cm}
% \label{wim_control_fig}
% \end{figure}

\subsubsection{WIM Interaction}
A primary goal of \ProjectName is to allow users to author precise paths to any destination in the environment shared with a robot. By incorporating semantic landmarks into the WIM, \ProjectName works toward this goal by ensuring authored paths are informed by meaningful features of the environment, such as obstacles standard navigation packages might not avoid, rather than arbitrary points in space. However, requiring users to perform highly granular authoring for each path, for example by manually placing every waypoint in a path, would be tedious and error-prone.

To achieve this goal and address this problem, the WIM allows users to select destinations and author paths through a drag-and-drop interaction. A user can place a 3D map pin at any desired location within the WIM, after which a series of blue waypoints is generated by the WIM controller from the robot’s current position to the selected pin, connected by a line as in Fig.~\ref{wim_control_fig}(b). These waypoints are not independent: each is linked to its neighbors through Unity spring joints (spring constant = 10,000; damping factor = 1,000). This design choice makes editing paths analogous to bending a wire—adjusting a single waypoint proportionally moves its neighbors—allowing users to shape paths quickly with only a few interactions. Translucent overlays of the waypoints are simultaneously rendered in the real environment, helping users anticipate how their authored path will translate to the physical space, seen in Fig.~\ref{wim_control_fig}(d). Once satisfied, users press a virtual button on their wrist to confirm, sending the destination and authored path to the edge server’s WIM generator and ultimately the Go2’s action planner. The robot then executes the navigation task, following the user-authored path to the selected destination.

\subsubsection{Go-To Commands}

\begin{figure}[t]
\centering
\vspace{0.15cm}
\includegraphics[width=\columnwidth]
{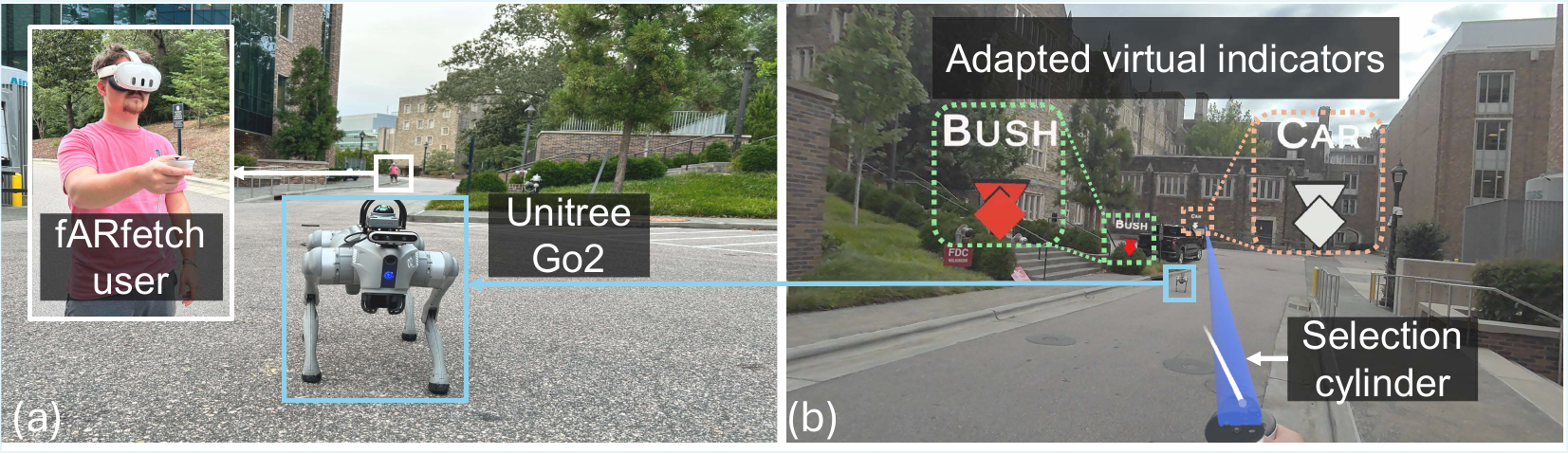}
\vspace{-0.5cm}
\caption{Example of fARfetch’s go-to command. (a) \ProjectName user with Go2 in visually diverse outdoor environment. (b) User selects distant, VLM-adapted real-world object indicator on car, sending location as destination for the Go2.}
\vspace{-0.5cm}
\label{go_to_figure}
\end{figure}

A challenge in large-scale AR-HRC is enabling users to issue contextually meaningful commands even when the robot or relevant landmarks are distant or occluded. However, users might not always want to stop and author a specific path to a destination with the WIM. To address this problem, \ProjectName implements a second type of command, which we call a ``go-to" command.

To do this, \ProjectName continuously synchronizes its WIM controller with the most up-to-date semantic map of the environment. This map integrates landmark names and poses detected by both the Quest and the Go2, expressed in their respective world and map frames. By sharing this semantic map with the virtual content controller, \ProjectName places labeled virtual indicators directly on detected real-world objects, as shown in Fig.~\ref{go_to_figure}(b) for a bush and car and Fig.~\ref{wim_control_fig}(c) for a wet floor sign. These real-world object indicators provide users with an enriched situational awareness, allowing them to perceive and reference environment landmarks even when distant or beyond VLOS.

To translate this awareness into control, we designed a pointing-style interaction inspired by prior work on distance-based selection in AR~\cite{VR_2025_force_paper}. When a user presses a button on the Quest controller, a long virtual cylinder appears (denoted the selection cylinder), enabling them to select an indicator at any distance, as seen in Fig.~\ref{go_to_figure}(b). When selected, an indicator is highlighted, and a subsequent button press issues a ``go-to" destination command to the edge server that is relayed to the robot's go-to listener and action planner. The robot then navigates to the landmark as in Fig.~\ref{go_to_figure}(b). Multiple indicators can be selected sequentially, allowing the user to author simple multi-destination paths. These interactions prioritize speed and intuitiveness for instances where a user needs to issue quick and simple commands to the robot.

\begin{figure}[t]
\centering
\vspace{0.15cm}
\includegraphics[width=\columnwidth]
{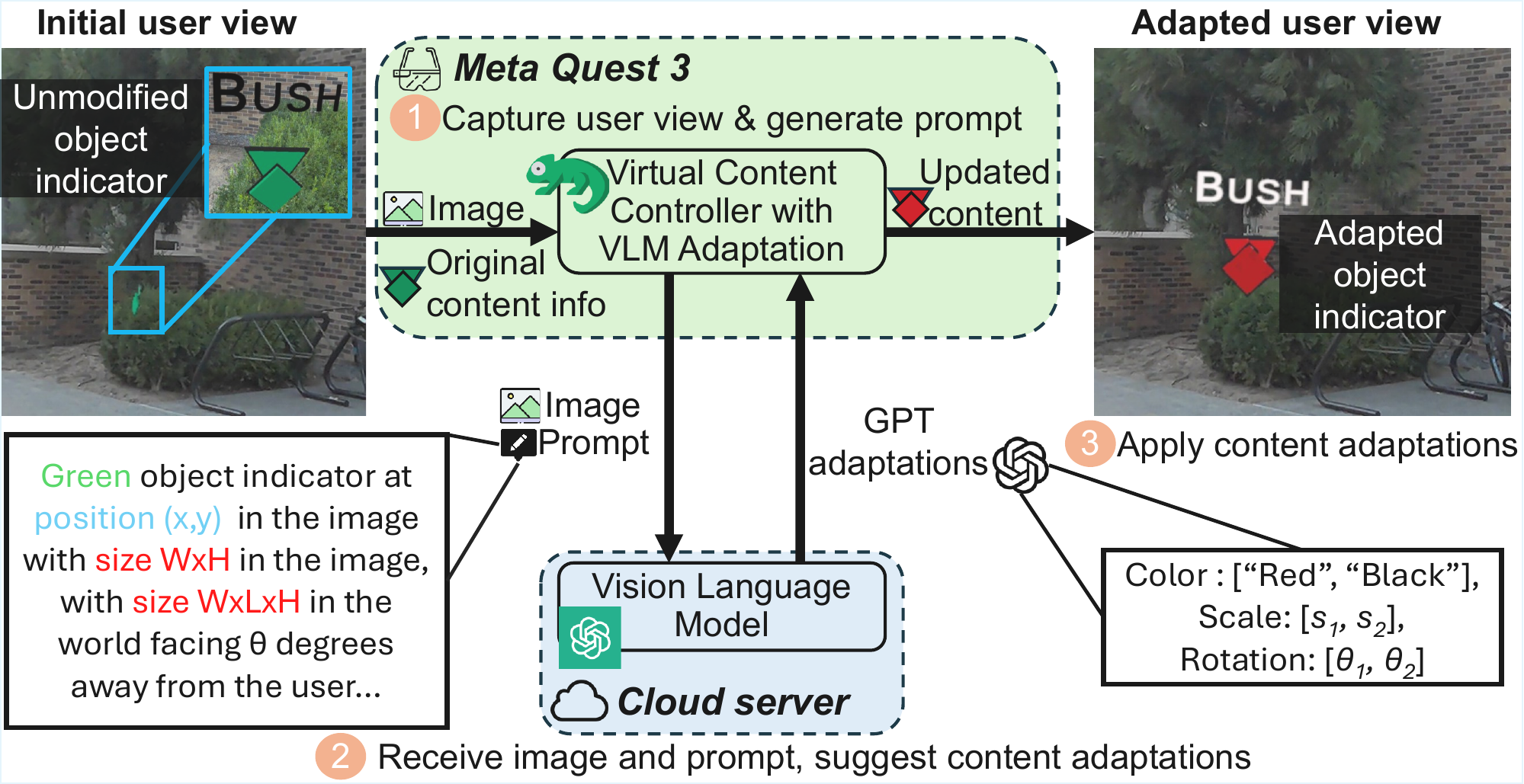}
\vspace{-0.5cm}
\caption{\ProjectName VLM-driven content adaptation process. Before adaptation, the bush indicator has poor contrast, is undersized, and faces away from the user; after adaptation, it has improved contrast, scale, and orientation.}
% \vspace{-0.5cm}
\label{content_adaptation_process}
\end{figure}

\begin{figure}[t]
\centering
\vspace{-0.0cm}
\includegraphics[width=\columnwidth]
{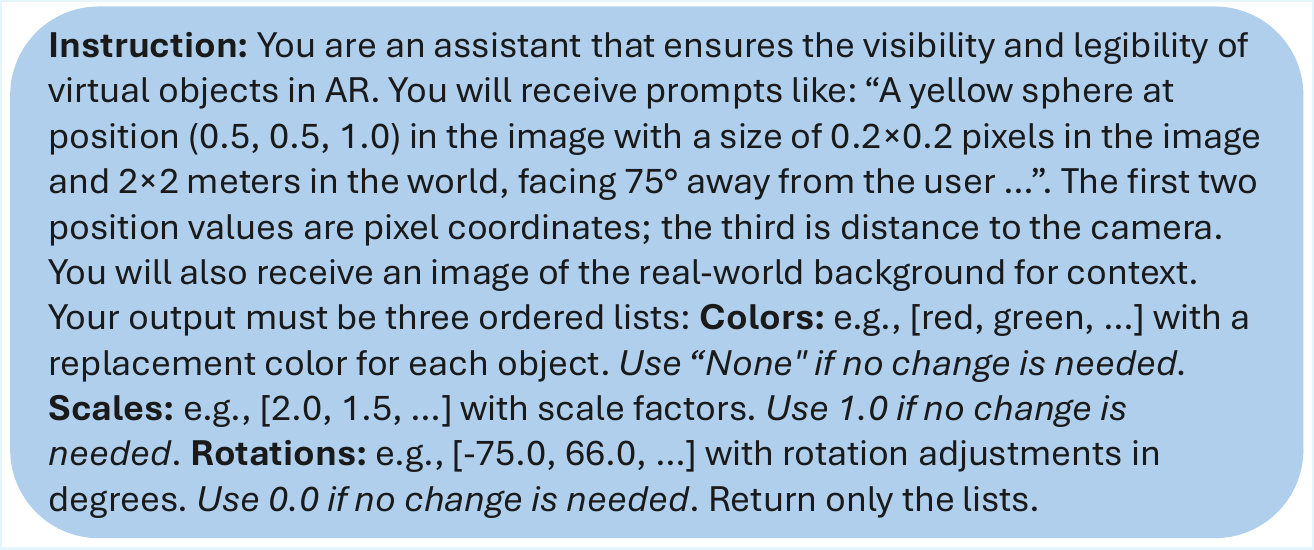}
\vspace{-0.5cm}
\caption{Instruction prompt used for AR content adaptation.}
% The instruction prompt given to a cloud VLM before content adaptation begins.
\vspace{-0.5cm}
\label{vlm_full_prompt}
\end{figure}

\subsection{VLM-Driven Content Adaptation}
Using a VLM, \ProjectName provides unified and scene-aware adaptation at runtime to maintain content legibility in diverse environments. Fig.~\ref{content_adaptation_process} shows an example of the process by which \ProjectName adapts virtual content for the case of a real-world object indicator displayed for a bush.

\subsubsection{Real-world Detection}
When new virtual content enters the Quest’s camera frustum, the virtual content controller triggers an adaptation cycle. It captures a raw image of the real-world scene without virtual content, denoted \( I \). At this time, there exists a set \( V = \{v_1, v_2, \cdots, v_n\} \) of virtual objects currently visible to the user. Capture is repeated any time virtual content enters the user's view. %allowing adaptation to occur throughout the entirety of \ProjectName use.
The captured images, together with the generated prompt, are transmitted to the VLM for generating content adaptation suggestions.

\subsubsection{Prompt Generation}
Once an image is captured, the system generates descriptive text strings for each visible virtual object. For each virtual object \( v_i \in V \), we define two complementary sets of parameters: its \emph{render state} and its \emph{perceptual context}. The render state specifies how the object is instantiated in AR:
\begin{equation}
    r_i = \{c_i, s_i^{w}, \phi_i^{w}\},
\end{equation}
where \( c_i \), \( s_i^{w} \), and \( \phi_i^{w} \) denote the object's color, world-space scale, and world-space orientation, respectively. The perceptual context captures how the object appears to the user relative to the real-world scene:
\begin{equation}
    q_i = \{p_i^{I}, s_i^{I}, d_i^{w}, \theta_i^{w}\},
\end{equation}
where \( p_i^{I} \) represents the object’s position in the image plane, \( s_i^{I} \) denotes its size in the image, \( d_i^{w} \) is its distance from the user in world space, and \( \theta_i^{w} \) captures its viewing angle relative to the user. These attributes describe the real-world scene context in which the object is perceived. These render states and perceptual contexts are used to construct structured descriptions for each object. %\emph{Color} (\( c_i \)) is used to evaluate contrast against the real-world background. \emph{Position} in the image plane (\( p_i^{I} \)), expressed in pixels, together with \emph{distance} from the Quest camera in world space (\( d_i^{w} \)), expressed in meters, informs the VLM of the object’s location in the user’s field of view and its corresponding background. \emph{Size}, expressed both in the image plane (\( s_i^{I} \)) and in world space (\( s_i^{w} \)), is used to assess legibility at the user’s viewing distance. \emph{Viewing angle} (\( \theta_i^{w} \)) indicates how the object is oriented relative to the user in world space. 
\emph{Color} is used to evaluate contrast against the real-world background. \emph{Position} in the image plane, together with \emph{distance} from the Quest camera in world space, informs the VLM of the object’s location in the user’s field of view and is used for color and scale reasoning. \emph{Size}, expressed both in the image plane and in world space, is also used for scale reasoning. \emph{Viewing angle} indicates how the object is oriented relative to the user in world space. The prompt is then created with explicit role assignment and formatting rules as shown in Fig.~\ref{vlm_full_prompt}, to guide the VLM toward consistent and effective adaptations through color choices, scale adjustments, and world-space rotation.

Rather than explicitly modeling the relationship between scene appearance and perceptual quality, \ProjectName leverages a VLM to infer this relationship. The VLM takes as input the image \( I \) and structured descriptions derived from \( \{r_i, q_i\} \), and outputs updated render states \( r_i' \) for each object.

\subsubsection{Adaptation Application}
The VLM returns adaptation types in separate comma-separated lists, with each item corresponding to a different virtual object. An example response is shown in Fig.~\ref{content_adaptation_process}. The virtual content controller receives the suggested adaptations and applies them to each object, producing an updated set of render states \( R' = \{r_1', r_2', \cdots, r_n'\} \), where each \( r_i' \) specifies the adapted color, world-space scale, and world-space orientation of object \( v_i \).

\section{Real-World Evaluation}
\label{sec:user_study}
% \vspace{-0.2cm}
\vspace{-0.22cm}
To evaluate \ProjectName, we conducted an IRB-approved within-subjects user study with 13 participants. In the study, participants wear the Meta Quest 3 and collaborate with the Unitree Go2 in order to complete a mock inspection of an environment, a common task in HRC scenarios~\cite{HRI_2019_Drone_Teleop_Indoors,CHI_2025_SafeSpect_AR_Drone_Building_Inspection,ISIC_2016_AR_Drone_Object_Structure_Inspection}. We design our study with the goal of evaluating three hypotheses: I) \ProjectName improves task efficiency and reduces the number and severity of mistakes made in a collocated HRC task in a large environment where VLOS is not guaranteed; II) \ProjectName reduces perceived workload in a collocated HRC task; and III) \ProjectName's VLM-driven content adaptation is effective at maintaining virtual content legibility in outdoor environments. Sec.~\ref{sec:study_design} describes the study task area and user task during the study. Sec.~\ref{sec:study_procedure_partic_selection} outlines the procedure for the study and the participant selection process. Sec.~\ref{sec:study_data_collection} details the survey and study performance data we collected.

% \begin{figure}[h!]
% \centering
% \vspace{-0.0cm}
% \includegraphics[width=\columnwidth]
% {figures/StudySetupFig.png}
% \vspace{-0.6cm}
% \caption{A top down view of the study task area and robot at the start of a study trial.}
% \vspace{-0.5cm}
% \label{study_setup_and_user_view}
% \end{figure}
\vspace{-0.2cm}
\subsection{Study Design}
\label{sec:study_design}

\noindent\textbf{Study Task.}
At the start of each trial, participants stand \textbf{\textit{30.5~m}} from the center of our designed task area, and remain at this location throughout the task. The robot is positioned at the edge of the task area, 1.5~m from its center, awaiting user commands.  Their objective is to direct the robot to visit all inspection targets in the task area as quickly as possible, in any order. A target is considered visited when the robot comes within the 0.5~m radius surrounding it. At the same time, participants are required to prevent the robot from entering any preset hazard areas within the task area.

\begin{figure*}[t]
	\centering
% \vspace{-2mm}
\hspace{-8mm}
\begin{minipage}{0.7\columnwidth}
\centering
\vspace{-0.0cm}
\includegraphics[width=\columnwidth]
{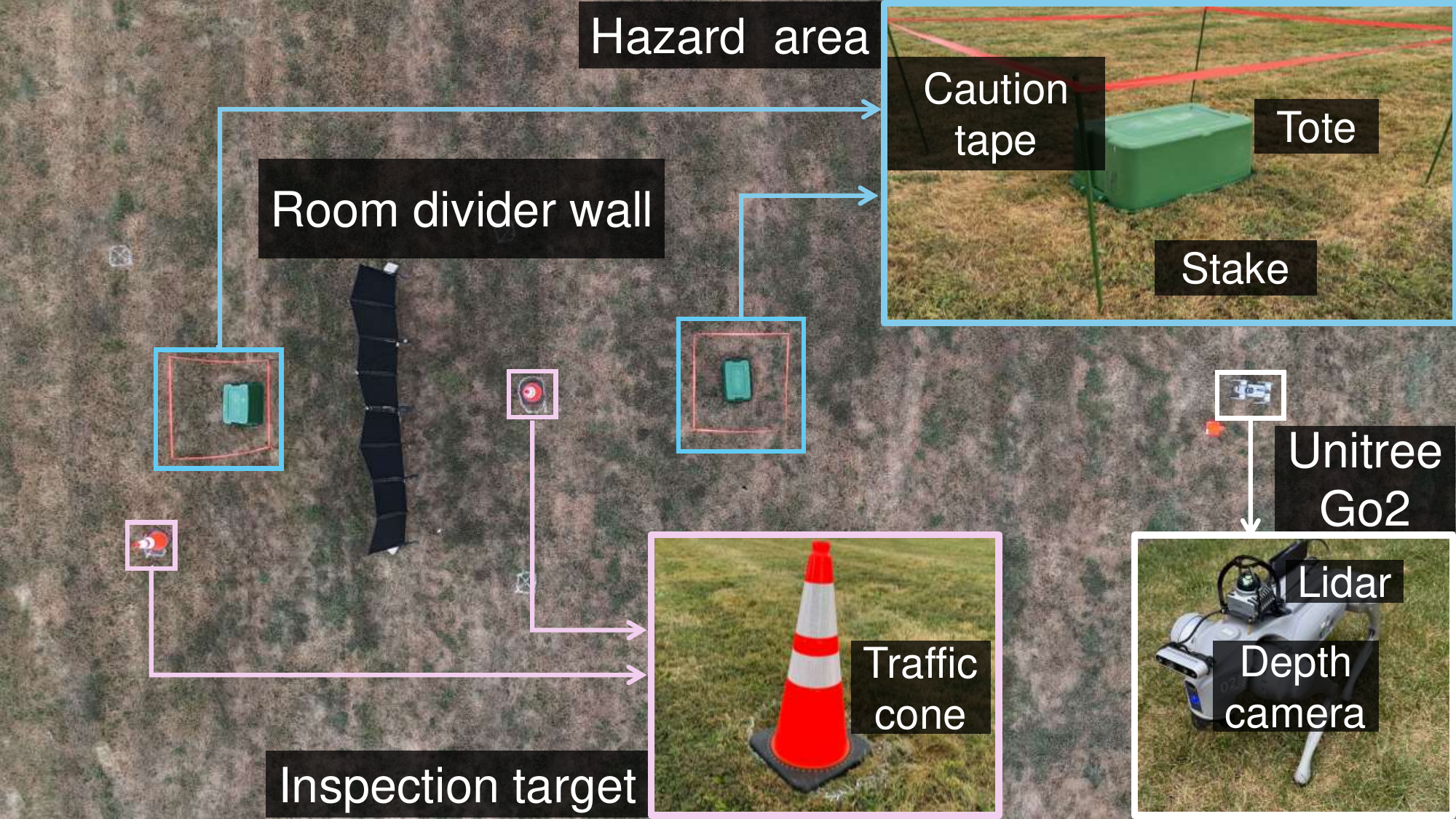}
\vspace{-0.55cm}
\caption{A top-down view of the study task area and robot at the start of a study trial.}
\vspace{-0.5cm}
\label{study_setup_and_user_view}
\end{minipage}
\hspace{.5mm}
\begin{minipage}{0.5\columnwidth}
\centering
\vspace{0.35cm}
\includegraphics[width=\columnwidth]
{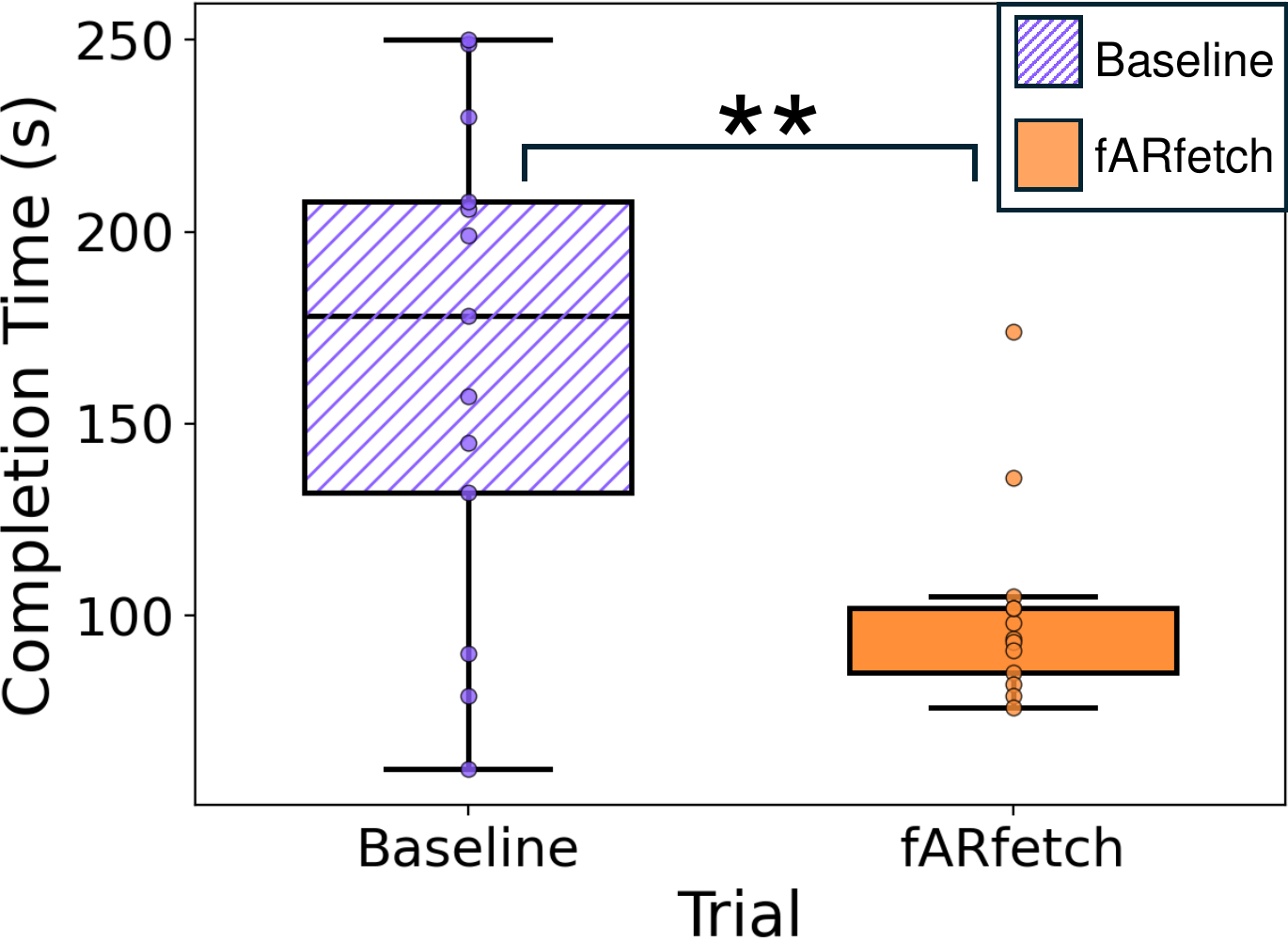}
\vspace{-0.5cm}
\caption{Task completion time results for all users in the baseline and AR trials. (**): $p\leq 0.01$}
\vspace{-0.5cm}
\label{study_time_results}
\end{minipage}
\hspace{.5mm}
\begin{minipage}{0.75\columnwidth}
\centering
\vspace{0.6cm}
\includegraphics[width=\columnwidth]
{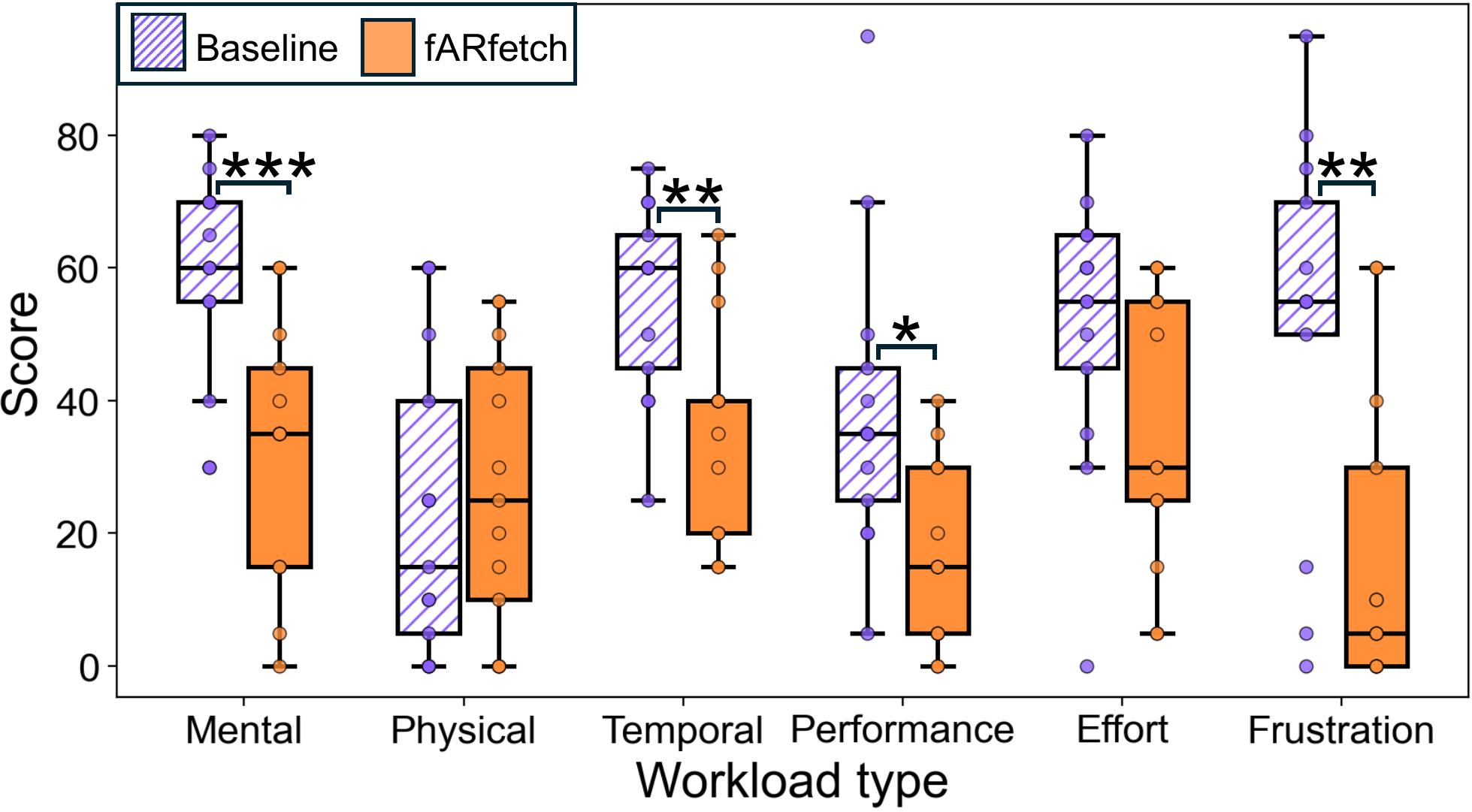}
\vspace{-0.5cm}
\caption{NASA-TLX results for all users in both trials of the study. (*): $p\leq 0.05$. (**): $p\leq 0.01$. (***): $p\leq 0.001$}
\vspace{-0.3cm}
\label{study_tlx_results}
\end{minipage}
\hspace{-10mm}
\vspace{-3mm}
\end{figure*}

\noindent\textbf{Task Area.}
Our task area is a square populated with three distinct types of objects: \emph{room dividers}, \emph{inspection targets}, and \emph{hazard areas}, as shown in Fig.~\ref{study_setup_and_user_view}. First, two 2.2×1.8~m \emph{room dividers} are positioned at the center of the space, aligned to form a wall bisecting the area. These dividers created conditions in which participants did not have direct VLOS to the robot. Second, traffic cones serve as \emph{inspection targets}, with a 0.5~m radius around each cone forming the valid inspection zone. We choose cones as the inspection targets due to their bright color and moderate size, making them easy for participants to see, and for the depth camera to get an accurate reading. One cone was placed on each side of the divider. Third, we designate \emph{hazard areas} by enclosing a tote within a 1.5×1.5~m square boundary formed by four stakes connected with caution tape. Entry is defined as any instance in which any part of the robot crosses the square boundary formed by stakes and caution tape, simulating regions the robot should avoid. This hazard design was chosen because the tape and stake boundaries present obstacles that are difficult for standard navigation and mapping systems to detect, while the size of the totes allows for accurate depth estimation by the depth camera. Together, these features emulate obstacles that might arise in a collocated human–robot collaboration task within a previously unseen environment. The layout of the targets and hazard areas are different on either side of the wall by design to create a more challenging task for participants.

% \noindent\textbf{Study Task.}
% At the start of each trial, the robot is positioned at the edge of the task area, 1.5~m from its center, awaiting user commands. Participants stand 30.5~m from the center of the area, facing one half of the space delineated by the room dividers, and remain at this location throughout the task. Their objective is to direct the robot to visit all inspection targets as quickly as possible, in any order. A target is considered visited when the robot comes within the 0.5~m radius surrounding it. At the same time, participants are required to prevent the robot from entering hazard areas, with entry defined as any instance in which part of the robot crosses the square boundary formed by the stakes and caution tape.

% When the task starts, the robot stands at the edge of the task area, awaiting commands from the user. The user stands at a distance of 15~m away from the center of the task area, facing one half of the area created by the room dividers; the user remains at this location for the duration of the task. During the task, the user must direct the robot to each inspection target within the space as quickly as possible. The inspection targets can be visited in any order the user chooses. An inspection target is considered visited if the robot stops within the 20~cm radius inspection area of the target. While directing the robot, the user must also work to avoid the hazard areas, with the robot considered as entering an area if any part of it crossed the square formed by the stakes and tape.

The task is performed in two trials: one using \ProjectName for robot control and one using RViz2~\cite{rviz} on a laptop with mouse and keyboard input. We selected RViz2 as the baseline because it provides a full-featured 2D map-based path planning interface built on the same SLAM and Nav2 stack as our AR condition, ensuring performance differences reflect interface modality rather than navigation capability. RViz2 has also been used as a baseline in prior AR-HRC work~\cite{ICRA_2021_ARROCH_AR_Robot_Collab_Indoors_TabletBased,ICRA_2024_MR_Robot_drag_and_drop_quadruped}, enabling comparison to existing literature. Unlike commercial mobile apps or joystick controllers, which often support only manual teleoperation or closed-source interfaces, RViz2 offers a standardized and extensible ROS2-based control framework.

% The task is performed in two trials: one using \ProjectName for robot control and one using Rviz2~\cite{rviz} on a laptop  with mouse and keyboard input for control. We selected RViz2 as our baseline for two primary reasons. First, RViz2 provides a full-featured 2D map-based path planning interface using the same SLAM and Nav2 stack as our AR condition, ensuring that differences in performance stem from interface modality rather than navigation capabilities. Second, it has been used as a baseline in prior AR-HRC work~\cite{ICRA_2021_ARROCH_AR_Robot_Collab_Indoors_TabletBased,ICRA_2024_MR_Robot_drag_and_drop_quadruped}, allowing comparison to existing literature. While commercial mobile apps or joystick controllers are common in practice, they typically provide either manual teleoperation without global path planning or closed-source interfaces that are not extensible across ROS2-based platforms. We therefore chose RViz2 as a standardized control interface.

In the \ProjectName trial, labeled virtual markers are placed at the real-world locations of inspection targets as \ProjectName detects each cone. Corresponding labeled markers also appear in the interface’s WIM representation of the environment. Each hazard area is likewise assigned a labeled virtual marker, distinct in appearance from those of the inspection targets, at both its real-world location and in the WIM. Participants direct the robot by issuing go-to commands or by selecting destinations and creating paths within the WIM, while avoiding the marked hazard areas. In the non-AR trial, participants control the robot using the RViz2 interface, where they could select destinations and create paths by manipulating individual 2D path segments similar to the AR interface. As in the \ProjectName trial, inspection targets and hazard areas are marked within the RViz2 interface.

% The task is performed in two trials: a trial using \ProjectName for robot control, and a trial using an Rviz-based non-AR baseline. During the \ProjectName trial, labeled virtual markers are placed at each inspection target's real-world location as \ProjectName detects each bucket, and are visible within \ProjectName's AR interface. Labeled markers are also placed within the interfaces WIM representation of the environment. Each hazard area also receives a labeled virtual marker (of a different appearance than those for the inspection targets) at both their real-world location and on the WIM representation. Users can then direct the robot through the task by using the go-to command technique (Fig.~\ref{study_setup_and_user_view}(b)) and by selecting destinations and creating paths for the robot using the WIM (Fig.~\ref{study_setup_and_user_view}(c)), working to avoid the marked hazard areas as they complete the task. In the non-AR trial, users can control the robot using a combination of an Rviz interface, with which they can select destinations and paths for the robot, and a joystick controller. The Rviz interface also displays the front camera view of the robot. As in the \ProjectName trial, the inspection targets and hazard areas are marked on the Rviz interface. 

%\vspace{-0.2cm}
\subsection{Study Procedure and Participant Selection}
\label{sec:study_procedure_partic_selection}
% \vspace{-1mm}
Upon arrival, participants read and signed a consent form and completed a pre-study survey. A researcher introduced the task and gave a tutorial on using \ProjectName and the RViz2 baseline to command the robot. For each interface, the researcher demonstrated how to issue robot movement commands, after which participants were allowed to try sending commands themselves before beginning the study trials. Each participant then completed two trials in a randomized order, completing the NASA Task Load Index (TLX) survey~\cite{NASATLX} following each trial. After completing the \ProjectName trial, participants completed a survey regarding the legibility of content during the trial. We recruited 13 participants from our university and the broader community (mean age: 25 years, range: 18--61; 46\% female). Participants reported varied prior AR headset experience: 2 used a headset more than once per week, 1 less than once per week, 8 had used one once or twice, and 2 had never used one. For robotic systems or RViz2, 6 participants had no prior experience, 6 had used them once or twice, and 1 used them frequently.

\subsection{Data Collection}
\label{sec:study_data_collection}
% \noindent\textbf{Pre-Study Test.}
% Prior to the study, participants completed the Perspective Taking/Spatial Orientation Test (PTSOT)\cite{Kozhevnikov2001SOT}, a validated measure of large-scale spatial ability. Performance on the PTSOT has been shown to correlate with both spatial layout learning and navigation performance\cite{Friedman2020DigitalSOT}. In the test, seven objects are displayed in a circular arrangement. For each of twelve items, participants are instructed to imagine themselves standing at one object, facing a second, and then indicate the direction to a third by drawing a line. Accuracy is quantified as the mean angular error across items, with lower error reflecting stronger spatial orientation ability. The test is administered over five minutes. In our analysis, we compare participants’ PTSOT scores with their performance in the study trials to evaluate hypothesis II, testing whether \ProjectName provides greater performance benefits for participants with lower spatial reasoning ability.
\noindent\textbf{Performance Data.}
We recorded the total time each participant required to complete the study task in each trial, measured from the initiation of robot commands to the arrival at the final inspection target. These completion times served as a measure of task efficiency. In addition, we documented the number of instances in which participants directed the robot into a hazard area, along with the duration of each incursion. Together, these measures captured both the frequency and severity of errors made during the task.

% \begin{table}[h!]
% \centering
% \caption{Virtual content visibility survey.} % Caption above the table
% % \resizebox{0.5\textwidth}{!}{
% \resizebox{\columnwidth}{!}{
% \begin{tabular}{@{}lp{8cm}@{}}
% \toprule
% \textbf{No.} & \textbf{Question} \\ 
% \midrule
% Q1 & The colors of the virtual elements allowed me to see them clearly against the real-world background. \\
% Q2 & The size of the virtual elements was appropriate for my viewing distance. \\
% Q3 & The orientation of the virtual elements relative to my viewpoint made them easy to read. \\
% \bottomrule
% \end{tabular}
% }
% \vspace{-0.0cm}
% \label{tab:vlm_survey}
% \end{table}

\noindent\textbf{Survey Data.}
Following both the \ProjectName and baseline trials, participants completed the NASA-TLX~\cite{NASATLX} to assess their perceived workload during each trial. After the \ProjectName trial, they also completed a custom legibility survey designed to capture their subjective impressions of the legibility of virtual content. The survey questions can be found in Fig.~\ref{study_viz_results}. These responses were used to evaluate the effectiveness of fARfetch’s VLM-adapted content. Each question was answered on a standard 5-point Likert scale.
\section{Results Analysis}
\label{sec:results}

This section presents the performance and survey results from the mock inspection task participants completed during our user study, and discusses their implications for the hypotheses presented in Sec.~\ref{sec:user_study}. For task completion times and NASA–TLX workload scores, we compared the baseline and \ProjectName trials using within‐subjects analyses. For each variable, difference scores were first evaluated for normality using the Shapiro–Wilk test. If the differences were approximately normally distributed, we applied a paired‐samples t-test. If normality was violated, we instead applied a Wilcoxon signed‐rank test. For all inferential tests, significance was assessed at $\alpha = 0.05$. %To analyze the relationship between participant performance data and TLX responses with participant performance on the PTSOT pre-test, we compute Pearson’s $r$  if the data is time or TLX data is normal, or Spearman’s $\rho$ if the data is non-normal.

% StudySetupFig_reducesizev3

% \begin{figure}[t]
% \centering
% \vspace{-0.0cm}
% \includegraphics[width=0.75\columnwidth]
% {figures/StudyTimeResults.png}
% \vspace{-0.3cm}
% \caption{Task completion time results for all users in the baseline and AR trials of the study. (**): $p\leq 0.01$}
% \vspace{-0.5cm}
% \label{study_time_results}
% \end{figure}

\noindent\textbf{Performance Results.}
Fig.~\ref{study_time_results} shows task completion times for the baseline and \ProjectName trials across all participants. A paired-samples t-test indicated a significant difference between trials, with \textit{participants completing the task 66\% faster in the \ProjectName trial compared to the baseline} (101.31~$s$ vs. 167.92~$s$, $p = 0.0013$). These findings suggest that \emph{\ProjectName substantially improved participant performance in a large-scale outdoor inspection task, supporting hypothesis I}. However, we observed no notable differences in error rates between conditions: only one participant entered a hazard area, doing so twice for a total duration of 39.53~s across both hazard area incursions. We note also that participants generally preferred to use WIM control when the robot was obscured, but otherwise preferred go-to commands.

% \begin{figure}[t]
% \centering
% \vspace{-0.0cm}
% \includegraphics[width=0.75\columnwidth]
% {figures/StudyTLXResults.png}
% \vspace{-0.3cm}
% \caption{NASA-TLX results for all users in both trials of the study. (*): $p\leq 0.05$. (**): $p\leq 0.01$. (***): $p\leq 0.001$}
% \vspace{-0.2cm}
% \label{study_tlx_results}
% \end{figure}
% \vspace{-2mm}
\noindent\textbf{NASA-TLX Results.}
Fig.~\ref{study_tlx_results} presents the NASA-TLX responses for all participants across both trials. Analysis of these responses revealed significant reductions in perceived workload across several workload types when using \ProjectName compared to the baseline. Specifically, mean mental demand scores were 43\% lower ($p = 0.0008$), temporal demand scores were 34\% lower ($p = 0.0087$), and frustration scores were 66\% lower ($p = 0.0042$) in the \ProjectName trial relative to baseline. These results suggest that \emph{\ProjectName substantially alleviated cognitive and temporal workload as well as frustration, supporting hypothesis II}. No significant differences were observed for physical demand ($p = 0.72$) or effort ($p = 0.079$). We posit that this reduction in workload stems from \ProjectName visualizing landmarks from the robot’s mapping directly into the user’s environment, eliminating the need for users to mentally reconcile separate representations of the space, and due to the intuitive path planning using the drag-and-drop interactions of the WIM with embedded landmarks. Interestingly, perceived performance scores were significantly higher in the baseline condition, with a 57\% higher mean rating compared to the \ProjectName trial ($p = 0.0114$). This suggests that although participants objectively performed better in the \ProjectName trial, they felt they had performed better in the baseline trial. We posit that this was due to the greater unfamiliarity of the AR interface compared to the 2D interface, leading to participants feeling as if they performed worse even when they did not.

\begin{figure*}[t]
\centering
% \vspace{0.2cm}
%\vspace{-2mm}
\includegraphics[width=1.4\columnwidth]
{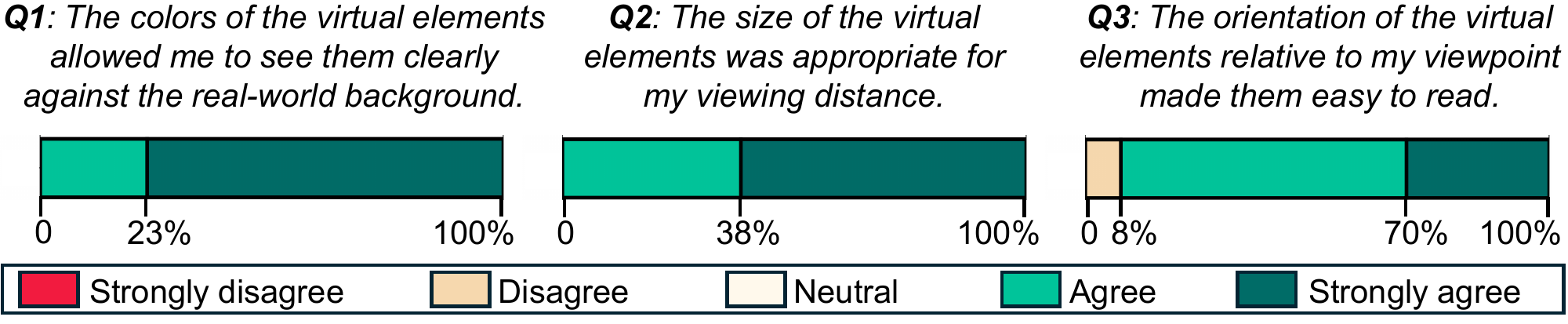}
% \vspace{-0.5cm}
\caption{\ProjectName virtual content legibility survey responses.}
\vspace{-0.5cm}
\label{study_viz_results}
\end{figure*}

% \vspace{-2mm}
\noindent\textbf{Legibility Survey.}
The results of our custom survey, designed to assess user perceptions of the legibility of virtual content managed by \ProjectName's VLM-driven method, are presented in Fig.~\ref{study_viz_results}. Across all questions, participants provided highly positive evaluations. For legibility based on color, all participants either agreed or strongly agreed, with most (76.9\%) selecting “strongly agree.” The appropriateness of content size was also rated favorably, with 61.5\% strongly agreeing and 38.5\% agreeing. Orientation received more moderate, yet still positive, responses: 92.3\% of participants agreed or strongly agreed, while one participant (7.7\%) disagreed. These findings \emph{support hypothesis III, indicating that fARfetch’s VLM-driven approach is effective at maintaining virtual content legibility in large outdoor environments, thereby improving the usability of AR for HRC}.

\section{Conclusion and Limitations}
\label{sec:conclusion}
This paper presents \ProjectName, an AR system supporting HRC in large, visually diverse environments where robot proximity and VLOS cannot be assumed. \ProjectName combines shared semantic mapping between robot and headset, a context-aware WIM for path authoring, landmark-grounded go-to commands, and VLM-driven view management for adapting content color, scale, and orientation. In a real-world inspection task, \ProjectName reduced completion time by 66\% and reduced mental demand, temporal demand, and frustration relative to a non-AR baseline, while maintaining strong content legibility. Our evaluation has several limitations. It assesses \ProjectName as an integrated system rather than isolating the contribution of each component, so future work should include ablations and component-level metrics for the AR interface, semantic landmark grounding, WIM path authoring, and VLM-driven adaptation. The VLM adaptation should also be evaluated with quantitative measures such as latency, frequency, and failure cases. In addition, RViz2 provides a standardized ROS-based baseline but is not optimized for collocated outdoor HRC, and additional training may reduce variability or improve baseline performance. Finally, the study used a modest sample size and a single inspection task, so future work should examine larger participant groups, additional tasks, more extensive training protocols, and more diverse environments.

\section*{ACKNOWLEDGMENT}
This work was supported in part by NSF grants CSR-2312760, CNS-2112562, and IIS-2231975, the National AI Institute for Edge Computing Leveraging Next Generation Wireless Networks Grant CNS-2112562, NSF CAREER Award IIS-2046072, NSF NAIAD Award 2332744, a Cisco Research Award, a Meta Research Award, Defense Advanced Research Projects Agency Young Faculty Award HR0011-24-1-0001, and the Army Research Laboratory under Cooperative Agreement Number W911NF-23-2-0224. The views and conclusions contained in this document are those of the authors and should not be interpreted as representing the official policies, either expressed or implied, of the Defense Advanced Research Projects Agency, the Army Research Laboratory, or the U.S. Government. This paper has been approved for public release; distribution is unlimited. No official endorsement should be inferred. The U.S.~Government is authorized to reproduce and distribute reprints for Government purposes notwithstanding any copyright notation herein. ChatGPT (OpenAI, San Francisco, CA, USA) was utilized to assist in drafting and refining sections of this manuscript. The authors have reviewed and approved all AI-assisted content to ensure accuracy and integrity.

%\bibliographystyle{IEEE_styles/IEEEtranMod}
%\bibliography{references.bib}
\printbibliography

\end{document}